\documentclass[]{TEAI}
\usepackage{helvet}

\usepackage{amsmath}
\usepackage{natbib}
\usepackage{graphicx}
\usepackage{subcaption}

\usepackage[toc,page,header]{appendix}
\usepackage[utf8]{inputenc}
\usepackage[T1]{fontenc}
\usepackage{hyperref}
\usepackage{url}
\usepackage{booktabs}
\usepackage{amsfonts}
\usepackage{nicefrac}
\usepackage{microtype}
\usepackage{xcolor}
\usepackage{wrapfig}
\usepackage{bm}
\usepackage{tikz}
\usepackage{array}
\usepackage{longtable}
\usepackage{pdflscape}
\usepackage{enumitem}
\usepackage{amssymb}
\usepackage{fontawesome}
\usepackage{titletoc}
\usepackage{comment}
\usepackage{tabularx}
\usepackage{minitoc}
\usepackage{etoolbox}

\definecolor{lightblue}{RGB}{200, 230, 255}
\definecolor{headerblue}{RGB}{150, 200, 255}
\definecolor{fwdbg}{RGB}{230,242,255}
\definecolor{fwdborder}{RGB}{120,165,220}
\definecolor{invbg}{RGB}{255,240,222}
\definecolor{invborder}{RGB}{220,160,90}

\usepackage{pgfplots}
\usepackage{float}
\usepackage{multirow}
\usepackage{makecell}
\usepackage{siunitx}
\usepackage[export]{adjustbox}
\usepackage{ragged2e}
\usepackage{caption}
\usepackage{pifont}
\usepackage[hang,flushmargin]{footmisc}
\usepackage{tcolorbox}
\tcbuselibrary{breakable}
\tcbuselibrary{skins}
\usepackage{listings}

\title{ThinkingVLA: Interleaved Vision and Language Reasoning for Robotic Manipulation}

\author{
    Tianyi Lu\textsuperscript{1,2},
    Hui Zhang\textsuperscript{1},
    Zijie Diao\textsuperscript{1},
    Junke Wang\textsuperscript{1},
    Shengqi Xu\textsuperscript{1},
    Xingyao Lin\textsuperscript{1,2},
    Guojin Zhong\textsuperscript{1},
    Ziyi Ye\textsuperscript{1},
    Peng Wang\textsuperscript{4},
    Zuxuan Wu\textsuperscript{1,2,3},
    Yu-Gang Jiang\textsuperscript{1},
}

\affiliation[1]{\mbox{Fudan University}}
\affiliation[2]{\mbox{Shanghai Innovation Institue}}
\affiliation[3]{\mbox{Neote AI}}
\affiliation[4]{\mbox{China Unicom}}

\abstract{
\begin{abstract}
Most Vision-Language-Action (VLA) models map observations directly to actions without explicit reasoning, limiting their capacity for reasoning-intensive long-horizon tasks.
To address this, existing approaches adopt Chain-of-Thought (CoT) reasoning to enable subgoal decomposition and spatial anticipation.
However, those methods lack a unified architecture for effective cross-modal reasoning and fail to explicitly include inverse reasoning ability based on the target state.
We argue that manipulation planning naturally decomposes into prediction, anticipating the next visual state, and inverse dynamics, inferring the actions to reach it.
Bridging both requires a unified autoregressive architecture that interleaves textual and visual reasoning in a single generation process.
We propose \textbf{ThinkingVLA}, a generative model that realizes this decomposition within a unified Mixture-of-Transformers architecture.
ThinkingVLA consists of a forward CoT that identifies the immediate subgoal and guides the visual forecasting; the predicted image then serves as the target state, grounding an inverse CoT that reasons about spatial relationships and action intent based on the predicted image; and the final action is generated conditioned on this full reasoning context.
Extensive experiments on simulation and real-world benchmarks demonstrate that ThinkingVLA consistently outperforms state-of-the-art baselines, with particularly large gains on long-horizon manipulation tasks.
\end{abstract}
}

\checkdata[Keywords]{Robotic Manipulation, Unified Model, Interleaved Chain-of-Thought}
\checkdata[Website]{\url{https://fitz135.github.io/thinkingvla.github.io/}}

\begin{document}
\maketitle

\begin{figure*}[t]
    \centering
    \includegraphics[width=\textwidth]{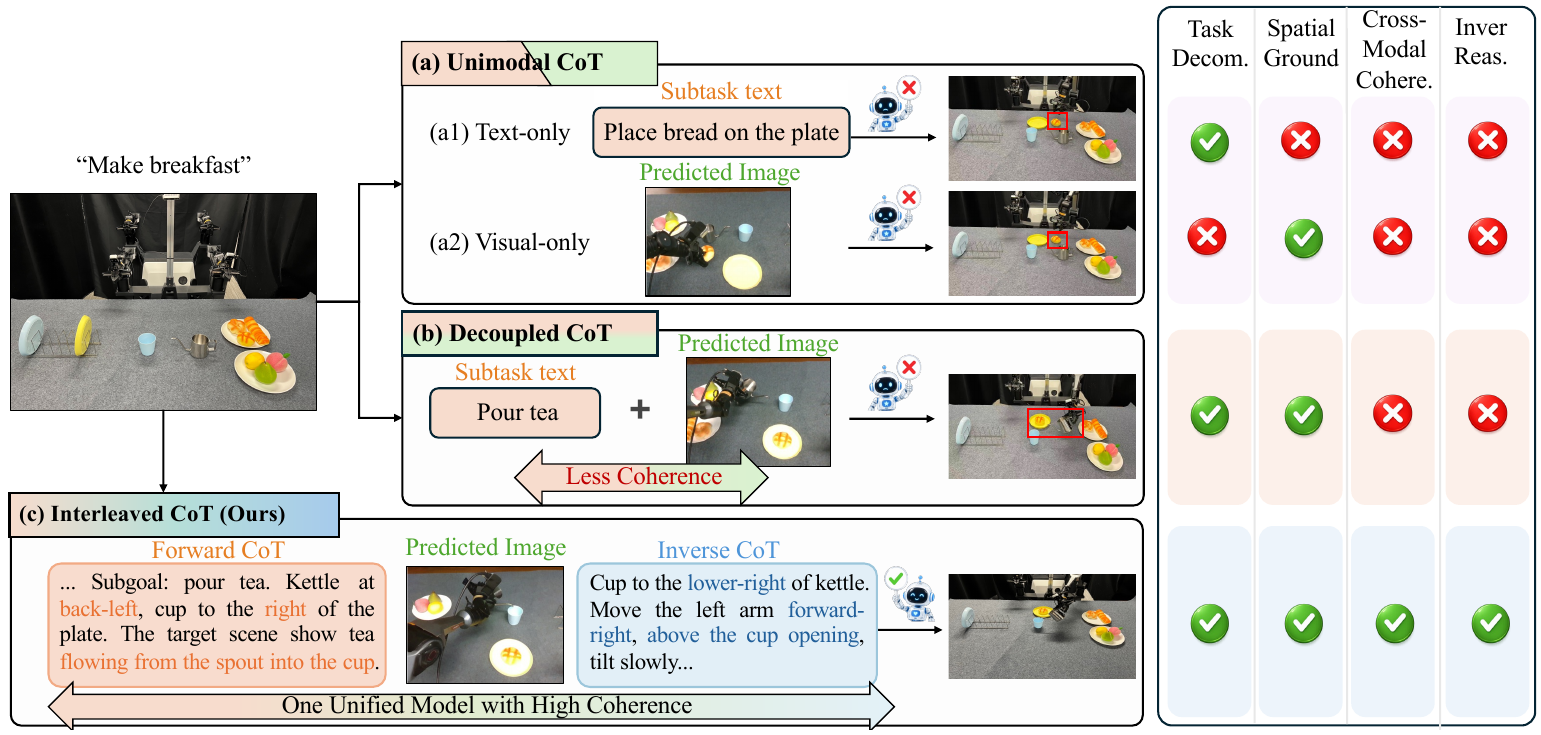}
    \caption{\textbf{Comparison of VLA reasoning strategies on a multi-step task.} We identify four capabilities required for structured manipulation reasoning: task decomposition, spatial grounding, cross-modal coherence, and inverse dynamics reasoning. \textbf{(a)}~Unimodal CoT provides at most one of the first two. \textbf{(b)}~Decoupled CoT adds both but lacks cross-modal coherence and inverse reasoning. \textbf{(c)}~ThinkingVLA achieves all four through an interleaved reasoning chain (forward CoT $\to$ predicted image $\to$ inverse CoT) within a single autoregressive sequence.}
    \label{fig:teaser}
\end{figure*}

\vspace{-3mm}
\section{Introduction}
\label{sec:intro}

Vision-Language-Action (VLA) models have made remarkable progress toward generalizable robot control by unifying visual perception, language understanding, and action generation within a single model~\citep{brohan2023rt,kimopenvla,black2024pi_0,black2025pi_}. 
Conventionally, VLA models compress the perception-to-action pipeline into a single implicit forward pass, where a vision-language backbone encodes the current observation and instruction and an action head directly decodes the motor command~\citep{brohan2023rt,kimopenvla,black2024pi_0,liurdt,wen2025diffusionvla,bjorck2025gr00t}. 
This monolithic mapping is effective on short-horizon, well-demonstrated tasks, but it provides no mechanism to decompose a complex instruction into subgoals, anticipate the consequences of a candidate action, or verify that a plan will achieve its intended effect. 
As tasks grow more complex (multi-step assembly, object rearrangement, long-horizon instruction following), a central question remains: how can we equip VLA models with the structured reasoning capacity that complex manipulation demands?

Drawing an analogy to the language domain, where Chain-of-Thought (CoT) prompting~\citep{wei2022chain} dramatically improves reasoning over direct answer generation, single-pass VLA models similarly lack the ``thinking'' capacity needed for challenging manipulation.
Recognizing this, recent work has introduced CoT reasoning into VLA models via two strategies. \emph{Unimodal CoT} augments the policy with reasoning in a single modality~\citep{zawalskirobotic,belkhale2024rt,black2025pi_,yin2025deepthinkvla,lee2025molmoact,zhao2025cot,cen2025worldvla,lv2025f1}: text-only approaches~\cite{zawalskirobotic,belkhale2024rt,black2025pi_,yin2025deepthinkvla,lee2025molmoact} improve multi-step performance but lack spatial precision, while visual-only approaches~\citep{tian2025predictive,zhao2025cot,cen2025worldvla,lv2025f1} provide spatial grounding but lack structured decomposition. \emph{Decoupled CoT}~\citep{zhang2025up,wang2025unified,ma2025unifying,gu2025manualvla,liu2026last,hu2026bagelvla} employs both modalities, but textual reasoning does not directly guide image generation, weakening cross-modal coherence, and no explicit reasoning follows the predicted image, leaving inverse dynamics implicit. 

Consider a robot instructed to ``make breakfast'' (Figure~\ref{fig:teaser}). It must retrieve a plate from the shelf and place it on the table, pick up bread and set it on the plate, then grasp the kettle and pour tea. At each step, the robot should anticipate what the scene will look like after the action, determine how to grasp and move each object, and verify that preceding steps were completed successfully before proceeding. This interplay between imagining future states and reasoning about how to reach them lies at the heart of manipulation planning. 

Motivated by this, we argue that two ingredients are needed. First, a principled decomposition: manipulation planning factors into prediction (anticipating the next visual state) and inverse dynamics (inferring the actions to reach it). Unimodal approaches address only one of these subproblems; decoupled approaches cover both but lack an explicit reasoning step that bridges the predicted future state to action generation, leaving inverse dynamics entirely implicit. The second ingredient is a unified autoregressive architecture in which text and image tokens share a common vocabulary: by interleaving textual reasoning, visual prediction, and visually-grounded inverse dynamics reasoning in a single unbroken causal sequence, the predicted image naturally serves as the bridge between prediction and inverse dynamics, with each step directly conditioning on all preceding outputs across modalities.

Building on this insight, we propose \textbf{ThinkingVLA}, a generative VLA model that realizes this decomposition within a unified Mixture-of-Transformers (MoT) architecture. ThinkingVLA interleaves textual reasoning and visual prediction in a single autoregressive pass: a forward CoT identifies the immediate subgoal and guides visual forecasting of the next state; the predicted image then serves as the bridge to inverse dynamics, grounding an inverse CoT that explicitly reasons about spatial relationships and action intent with precision that language alone cannot provide; and the final action is generated conditioned on this full reasoning context. The entire model is end-to-end trainable, and every intermediate output remains human-inspectable. We adopt a progressive training pipeline analogous to pretraining, continual pretraining, and fine-tuning in large language models: ThinkingVLA first acquires general-purpose reasoning and visual prediction capabilities, then learns end-to-end action generation across diverse manipulation domains, and finally adapts to each target robot embodiment, yielding a reasoning foundation model for manipulation whose structured multimodal reasoning transfers across robot platforms and task distributions.

Our main contributions are as follows:
\begin{itemize}[leftmargin=*]
    \item We identify the decomposition of manipulation planning into prediction and inverse dynamics as a principled framework for VLA reasoning, and introduce ThinkingVLA, a unified autoregressive model that bridges the two by interleaving language subgoal identification, future image prediction, and visually-grounded action reasoning within a single coherent generation process.
    \item We design a progressive training pipeline, from generative reasoning pretraining to end-to-end policy learning, that enables effective acquisition of structured multimodal reasoning capabilities, with pretraining substantially improving data efficiency on downstream manipulation tasks.
    \item We demonstrate through extensive simulation and real-world experiments that ThinkingVLA outperforms both unimodal and decoupled CoT approaches, with ablations confirming the contribution of each reasoning stage.
\end{itemize}

\section{Related Work}
\label{related_work}

\paragraph{Vision-Language-Action Models.}
Vision-Language-Action (VLA) models unify visual perception, language understanding, and action generation within a single model, and have demonstrated strong generalization when pretrained on large, diverse datasets~\citep{brohan2023rt,belkhale2024rt,kimopenvla,liurdt,lin2026activemimic}. A common architectural pattern pairs a VLM backbone with a dedicated action expert, using flow matching~\citep{black2024pi_0,bjorck2025gr00t,black2025pi_} or diffusion~\citep{wen2025diffusionvla,wen2025tinyvla}, to produce smooth, continuous trajectories.

Recent work augments VLA policies with Chain-of-Thought (CoT) reasoning~\citep{wei2022chain}. Unimodal text-only methods prepend intermediate language plans, such as step-by-step instructions, or language subgoals, before action generation~\citep{zawalskirobotic,belkhale2024rt,black2025pi_,lee2025molmoact,yin2025deepthinkvla}. Unimodal visual methods generate future observation states or visual subgoal representations as intermediate outputs~\citep{tian2025predictive,zhao2025cot,cen2025worldvla,lv2025f1}. Decoupled multimodal methods, including UP-VLA~\citep{zhang2025up}, UniVLA~\citep{wang2025unified},VITA~\citep{ma2025unifying}, ManualVLA~\citep{gu2025manualvla}, LaST$_{0} $~\citep{liu2026last}, and BagelVLA~\citep{hu2026bagelvla}, incorporate both textual and visual reasoning as separate sequential stages. In contrast, ThinkingVLA interleaves both modalities within a single autoregressive pass, with each reasoning step directly conditioning on all preceding outputs across modalities.

\paragraph{Unified Vision-Language Generation.}
A key enabler of ThinkingVLA is the ability to generate both text and images within a single autoregressive model. This capability builds on the development of discrete image tokenizers which compress images into sequences of codebook indices, allowing image generation to be cast as next-token prediction over a shared vocabulary with text~\citep{van2017neural, razavi2019generating, esser2021taming, yulanguage, wang2024omnitokenizer, shi2025scalable}. Building on this foundation, a growing body of work has explored unified multimodal models that interleave text and image tokens in a single sequence for joint understanding and generation~\citep{team2024chameleon, wang2024emu3, zhoutransfusion, xieshow, cui2025emu3, deng2025emerging, wu2025janus, chen2025janus, wuvila, wu2026liquid, maunitok}. Among these, one line of work adopts a purely token-based architecture that handles all modalities with a unified autoregressive transformer~\citep{team2024chameleon,wang2024emu3, wu2025janus, chen2025janus}, while another integrates next-token prediction for text with diffusion processes for images within a single model~\citep{zhoutransfusion,xieshow}. ThinkingVLA follows the token-based autoregressive route so that each token directly conditions on all preceding outputs across modalities, and repurposes this capability for robotic planning: generated image tokens serve as functional intermediates bridging language reasoning and action generation within a closed-loop pipeline.

\section{Method}
\label{method}

\subsection{Task Formulation}

We formulate instruction-conditioned robotic manipulation as a sequential decision-making problem. At each decision step $t$, the robot receives a task instruction $l$ and the visual observation $o_t$, and predicts an action chunk $a_t$. Standard VLA policies model this as a direct conditional mapping $p(a_t \mid o_t, l)$ \citep{brohan2023rt,kimopenvla,black2024pi_0}, which leaves the intermediate planning process entirely implicit. ThinkingVLA instead makes this planning process explicit by introducing three structured intermediate variables at each step: a forward CoT $r_t^{\mathrm{fwd}}$ that interprets the task instruction and reasons about the desired next state, one or more predicted future images $\hat{o}_{t+1}$ that visualize that state from the viewpoints available on the robot, guided by the forward reasoning, and an inverse CoT $r_t^{\mathrm{inv}}$ that reasons about how to reach it through concrete actions. We therefore factorize action generation as
\begin{equation}
\begin{aligned}
p_{\theta}(&r_t^{\mathrm{fwd}}, \hat{o}_{t+1}, r_t^{\mathrm{inv}}, a_t \mid o_t, l) \\
&= \underbrace{p_{\theta}(r_t^{\mathrm{fwd}} \mid o_t, l)\;
p_{\theta}(\hat{o}_{t+1} \mid o_t, l, r_t^{\mathrm{fwd}})}_{\text{prediction}} \\
&\quad\;\times\;
\underbrace{p_{\theta}(r_t^{\mathrm{inv}} \mid o_t, l, r_t^{\mathrm{fwd}}, \hat{o}_{t+1})\;
p_{\theta}(a_t \mid o_t, l, r_t^{\mathrm{fwd}}, \hat{o}_{t+1}, r_t^{\mathrm{inv}})}_{\text{inverse dynamics}}.
\end{aligned}
\label{eq:task_formulation}
\end{equation}

We realize this factorization with a Mixture-of-Transformers (MoT) architecture in which a thinking expert generates the interleaved reasoning sequence $[r_t^{\mathrm{fwd}},\; \hat{o}_{t+1},\; r_t^{\mathrm{inv}}]$, and a dedicated action expert receives the full reasoning context to produce $a_t$. The full pipeline remains end-to-end trainable. We detail the architecture and training procedure in \S\ref{sec:architecture} and \S\ref{sec:training}, respectively.

\subsection{Architecture}
\label{sec:architecture}

\begin{figure}[t]
    \centering
    \includegraphics[width=\linewidth]{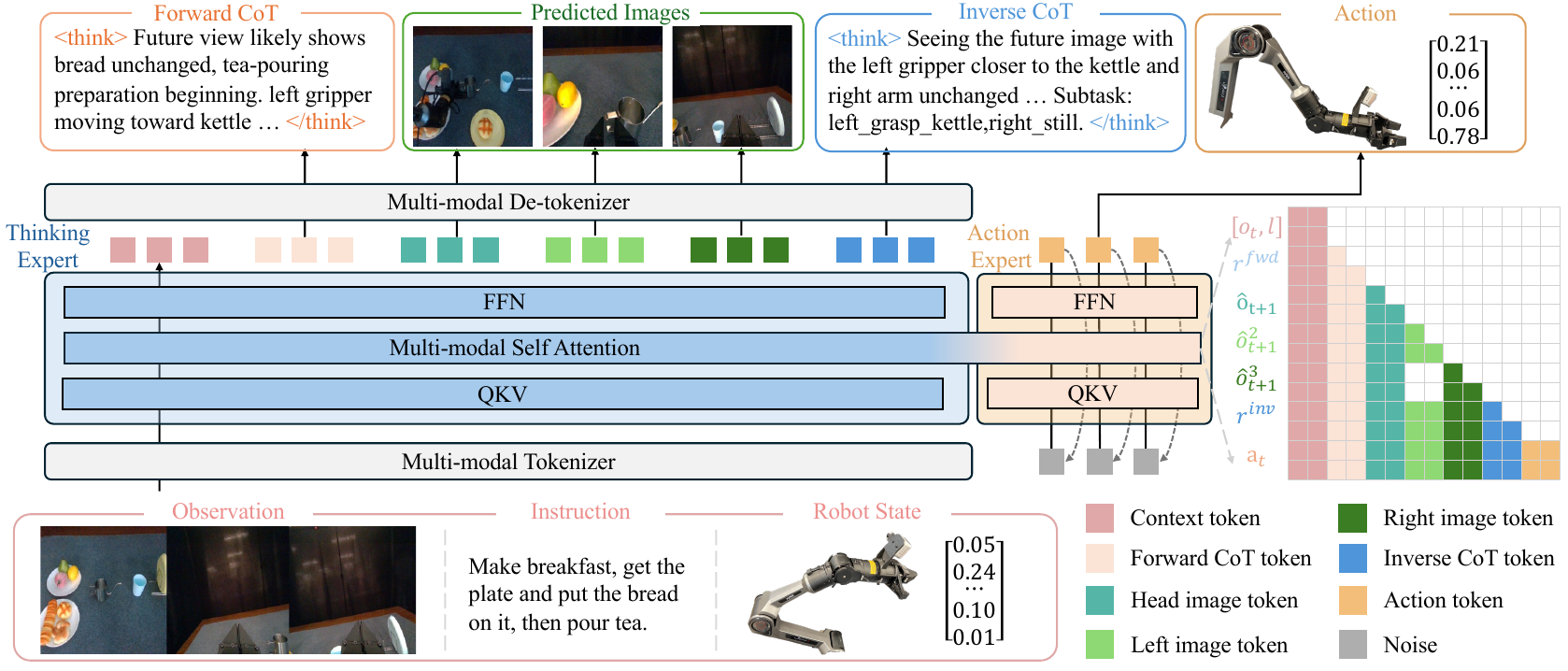}
    \caption{Overview of the ThinkingVLA architecture. The thinking expert generates interleaved textual reasoning and future image tokens; the action expert produces the action chunk via flow matching. The attention mask (right) illustrates token-level visibility in the multi-view setting.}
    \label{fig:pipeline}
\end{figure}

ThinkingVLA adopts a Mixture-of-Transformers (MoT) architecture with two coupled experts (Figure~\ref{fig:pipeline}): a thinking expert that generates the reasoning chain $[r_t^{\mathrm{fwd}},\;\hat{o}_{t+1},\;r_t^{\mathrm{inv}}]$ for the prediction phase of Eq.~\eqref{eq:task_formulation}, and an action expert that produces the action chunk $a_t$ for inverse dynamics. 

\paragraph{Unified tokenization.}
A multi-modal tokenizer maps all inputs and outputs into a shared token space: language is tokenized with the pretrained VLM~\citep{beyer2024paligemma} text tokenizer, observations are encoded via a SigLIP vision encoder~\citep{zhai2023sigmoid}, and images are discretized into vision tokens with a Cosmos image tokenizer~\citep{agarwal2025cosmos}. Text and vision tokens share a unified vocabulary, so the thinking expert handles both modalities in a single autoregressive pass. The training sequence takes the form:
\begin{equation}
    [\texttt{BOS}]
    \underbrace{\;\{o_t\}\;\{l\}}_{\text{context}}\;\underbrace{[\texttt{think}]\;\{r_t^{\mathrm{fwd}}\}\;[\texttt{/think}]\;[\texttt{gen}]\;\{\hat{o}_{t+1}\}\;[\texttt{/gen}]\;[\texttt{think}]\;\{r_t^{\mathrm{inv}}\}\;[\texttt{/think}]}_{\text{reasoning chain}}
    [\texttt{EOS}]
    \label{eq:token_sequence}
\end{equation}
The context consists of observation tokens $\{o_t\}$ (ViT features concatenated with Cosmos discrete tokens), instruction tokens $\{l\}$, and robot proprioceptive state encoded alongside the instruction. In the reasoning chain, special tokens \texttt{[think]}/\texttt{[/think]} bracket textual CoT segments and \texttt{[gen]}/\texttt{[/gen]} bracket the predicted future image, serving as modality-switching signals. Because text and image tokens share the same vocabulary and are generated by the same autoregressive process, the reasoning chain $[r_t^{\mathrm{fwd}},\;\hat{o}_{t+1},\;r_t^{\mathrm{inv}}]$ forms a single unbroken causal sequence with no modality boundary: (1)~the image generation loss propagates directly through the preceding forward CoT tokens, so the model learns to produce text that facilitates accurate visual prediction---a gradient coupling that requires no separate cross-modal interface; (2)~the subsequent inverse CoT $r_t^{\mathrm{inv}}$ conditions on individual image tokens through the same causal attention used for text, grounding action reasoning in the spatial details of the predicted scene without any cross-modal projection.

\paragraph{Mixture-of-Transformers.}
Both experts are multi-layer transformers with separate parameters that share attention at every layer: each expert computes its own Q/K/V projections; the K/V pairs are concatenated and a single joint attention is applied; the output is then split back and passed through expert-specific feed-forward networks.

As illustrated in Figure~\ref{fig:pipeline}, the attention mask adaptively mixes bidirectional and causal patterns: context tokens attend bidirectionally; reasoning and image tokens are generated causally; and the action expert attends to the full thinking prefix while remaining invisible to it, enabling it to leverage the reasoning context without disrupting generation. For multi-view prediction, each wrist-view image attends to all third-view tokens for spatial consistency but not to other wrist views.

\paragraph{Training objective.}
The thinking expert is supervised by next-token prediction over the reasoning chain in Eq.~\eqref{eq:token_sequence}. We decompose this into two terms according to token modality. The text loss supervises CoT generation:
\begin{equation}
    \mathcal{L}_{\text{text}} = -\sum_{i \in \mathcal{T}} \log p_\theta\!\left(y_i \mid y_{<i},\, o_t,\, l\right),
    \label{eq:text_loss}
\end{equation}
where $\mathcal{T}$ indexes the text token positions—forward CoT and inverse CoT; note that the latter is autoregressively conditioned on the preceding image tokens $\hat{o}_{t+1}$. The image loss supervises future-image generation:
\begin{equation}
    \mathcal{L}_{\text{img}} = -\sum_{j \in \mathcal{V}} \log p_\theta\!\left(y_j \mid y_{<j},\, o_t,\, l\right),
    \label{eq:img_loss}
\end{equation}
where $\mathcal{V}$ indexes all vision token positions in the reasoning chain; in single-view prediction $\mathcal{V}$ corresponds to $\hat{o}_{t+1}$, while in multi-view prediction it spans all generated views. Both losses are standard cross-entropy computed within the thinking expert.

The action expert is supervised by a flow-matching loss. Given time index $\tau \in [0,1]$ and Gaussian noise $\epsilon$, a perturbed action $x_\tau = \tau\,\epsilon + (1-\tau)\,a_t$ is constructed, and the expert learns to predict the target velocity $u = \epsilon - a_t$ conditioned on $x_\tau$, $\tau$, and the reasoning context $c_t = [r_t^{\mathrm{fwd}},\,\hat{o}_{t+1},\,r_t^{\mathrm{inv}}]$ provided by the thinking expert via shared attention:
\begin{equation}
    \mathcal{L}_{\text{action}} = \mathbb{E}_{\tau,\,\epsilon}\!\left[\left\| v_\theta(x_\tau,\,\tau,\,c_t) - u \right\|^2\right].
    \label{eq:action_loss}
\end{equation}
The full training objective combines the three losses:
\begin{equation}
    \mathcal{L} = \lambda_{\text{text}}\,\mathcal{L}_{\text{text}} + \lambda_{\text{img}}\,\mathcal{L}_{\text{img}} + \lambda_{\text{action}}\,\mathcal{L}_{\text{action}},
    \label{eq:total_loss}
\end{equation}
where $\lambda_{\text{text}}$, $\lambda_{\text{img}}$, and $\lambda_{\text{action}}$ are per-stage loss weights specified in \S\ref{sec:training}.

\subsection{Training and Inference Strategy}
\label{sec:training}

Because ThinkingVLA combines reasoning, visual prediction, and action generation---each requiring distinct supervision signals---we adopt a multi-stage training strategy that introduces these capabilities progressively: reasoning and visual prediction (Stage~1), end-to-end action generation (Stage~2), and target robot adaptation (Stage~3).

\paragraph{Training data.}
We train on a mixture of robotic manipulation datasets: Stages~1--2 use a subset of the Open X-Embodiment (OXE) dataset~\citep{open_x_embodiment_rt_x_2024}, while Stage~3 fine-tunes on RoboTwin~\citep{chen2025robotwin} and real-world demonstrations. For reasoning supervision, we generate forward CoT $r_t^{\mathrm{fwd}}$, inverse CoT $r_t^{\mathrm{inv}}$ by prompting Vision-Language Models. Dataset compositions and the annotation pipeline are detailed in Appendix~\ref{app:dataset} and \ref{app:thinking}.

\paragraph{Stage 1: Generative reasoning pretraining.}
We train only the thinking expert to generate forward reasoning and a single future image, without action prediction. Given a demonstration trajectory, the model receives the observation $o_t$ and instruction $l$, and is supervised to produce the forward CoT $r_t^{\mathrm{fwd}}$ and the predicted future image $\hat{o}_{t+1}$. The loss weights are $\lambda_{\text{text}} = \lambda_{\text{img}} = 1$ with $\lambda_{\text{action}} = 0$.

\paragraph{Stage 2: End-to-end policy learning.}
We introduce the action expert and enable the complete pipeline. The model now generates the full reasoning chain $[r_t^{\mathrm{fwd}},\,\hat{o}_{t+1},\,r_t^{\mathrm{inv}}]$ and predicts actions $a_t$ under the full objective in Eq.~\eqref{eq:total_loss} with $\lambda_{\text{text}} = 1$, $\lambda_{\text{img}} = 2$, and $\lambda_{\text{action}} = 10$.

\paragraph{Stage 3: Target robot adaptation.}
We fine-tune the full model on each target dataset. To match each dataset's camera configuration, single-view future image generation is extended to multi-view prediction. The loss weights are $\lambda_{\text{text}} = \lambda_{\text{img}} = 1$ and $\lambda_{\text{action}} = 10$.

\paragraph{Inference strategy.}
We employ two complementary techniques to balance generation quality and inference efficiency; training-time implementation details are provided in Appendix~\ref{app:cfg}.

\emph{Classifier-free guidance (CFG)}~\citep{ho2021classifier} improves future-image prediction quality. At inference, we fuse logits from three condition-ablated branches:
\begin{equation}
    l_{\text{cfg}} = l_{\text{base}} + s_I\,(l_{\text{img}} - l_{\text{base}}) + s_T\,(l_{\text{full}} - l_{\text{img}}),
    \label{eq:cfg}
\end{equation}
where $l_{\text{full}}$, $l_{\text{img}}$, and $l_{\text{base}}$ are logits under full, image-only, and unconditional conditioning, and $s_I$, $s_T$ control guidance strength. CFG is applied exclusively to future-image token generation; reasoning and action prediction are not guided.

\emph{Reasoning dropout} enables flexible efficiency--quality trade-offs. By randomly dropping $r_t^{\mathrm{fwd}}$ and $r_t^{\mathrm{inv}}$ independently during training, the model learns to generate and predict even without explicit reasoning, allowing selective skipping of CoT generation at inference to reduce latency.

\section{Experiments}
\label{exp}

We evaluate ThinkingVLA in simulation (\S\ref{sec:sim}) and real-world settings (\S\ref{sec:realworld}), and conduct ablation studies on real-world tasks (\S\ref{sec:ablation}) to answer three questions:
\textbf{(1)}~Does ThinkingVLA outperform state-of-the-art VLA baselines, and does the advantage grow with task horizon?
\textbf{(2)}~How does each component of the interleaved reasoning chain contribute to task performance?
\textbf{(3)}~How critical is large-scale end-to-end action pretraining for downstream adaptation?

\subsection{Experimental Setup}
\label{sec:setup}

ThinkingVLA builds on the $\pi_{0.5}$~\citep{black2025pi_} architecture. The thinking expert uses PaliGemma~\citep{beyer2024paligemma} as its backbone, with a Cosmos image tokenizer~\citep{agarwal2025cosmos} for discrete visual tokenization. The action expert is a 300M-parameter transformer that shares attention with the thinking expert at every layer. We train the model following the three-stage procedure described in \S\ref{sec:training}. Additional architecture and training details are provided in Appendix~\ref{app:architecture} and~\ref{app:training}.

\paragraph{Baselines.}
We compare against five representative VLA models:
$\bm{{\pi}_0}$~\citep{black2024pi_0}, a flow-matching VLA with a dedicated action expert;
$\bm{{\pi}_{0.5}}$~\citep{black2025pi_}, a VLA model with subtask prediction;
\textbf{XVLA}~\citep{zheng2025x}, a scalable cross-embodiment VLA model;
\textbf{UP-VLA}~\citep{zhang2025up}, a unified understanding and prediction model;
and \textbf{BagelVLA}~\citep{hu2026bagelvla}, a unified model that integrates linguistic planning, visual forecasting, and action generation.
For each baseline, we evaluate using officially released checkpoints when available and cite published results otherwise.
In simulation, we compare against $\pi_0$, XVLA, UP-VLA, and BagelVLA (\S\ref{sec:sim}); in real-world experiments, we compare against $\pi_0$, $\pi_{0.5}$, and UP-VLA (\S\ref{sec:realworld}).

\subsection{Evaluation in Simulation}
\label{sec:sim}

\paragraph{RoboTwin 2.0.}
RoboTwin 2.0~\citep{chen2025robotwin} is a dual-arm manipulation benchmark with diverse tasks spanning object manipulation, tool use, and coordinated bimanual operations.
Following the horizon-based task grouping of Lingbot-VA~\citep{li2026causal}, we select 20 RoboTwin tasks: 10 at Horizon$=$1, 5 at Horizon$=$2, and 5 at Horizon$=$3. We used the official settings: for each task, we used 50 episodes for fine-tuning, and evaluated each task 100 times under both the Easy and Hard settings.

\paragraph{Results.}
Table~\ref{tab:robotwin} summarizes the simulation results.
ThinkingVLA achieves the highest average success rate on Easy (77.9\%) and competitive performance on Hard (29.3\%). The advantage grows with task horizon: against BagelVLA, which also generates future images and subtask text, ThinkingVLA improves by 15.2\,pp on Horizon$=$2 Easy and 19.6\,pp on Hard; on Horizon$=$3, ThinkingVLA reaches 60.0\% Easy / 23.6\% Hard versus 55.8\% / 6.8\% for BagelVLA and 44.6\% / 18.6\% for XVLA, indicating that interleaved reasoning provides tighter cross-modal coupling than independently generating text and image. Per-task results are provided in Appendix~\ref{app:robotwin}.

\begin{table*}[t]
  \centering
  \caption{Results on RoboTwin 2.0. We report success rate (\%) on Easy and Hard splits across task horizons. Best results are \textbf{bold}; second-best are \underline{underlined}.}
  \label{tab:robotwin}
  \setlength{\tabcolsep}{5pt}
  \resizebox{\textwidth}{!}{%
  \begin{tabular}{l cc cc cc cc cc}
    \toprule
    & \multicolumn{2}{c}{$\pi_0$~\citep{black2024pi_0}}
    & \multicolumn{2}{c}{XVLA~\citep{zheng2025x}}
    & \multicolumn{2}{c}{UP-VLA~\citep{zhang2025up}}
    & \multicolumn{2}{c}{BagelVLA~\citep{hu2026bagelvla}}
    & \multicolumn{2}{c}{\textbf{ThinkingVLA (Ours)}} \\
    \cmidrule(lr){2-3}\cmidrule(lr){4-5}\cmidrule(lr){6-7}\cmidrule(lr){8-9}\cmidrule(lr){10-11}
    \textbf{Metric}
      & Easy & Hard & Easy & Hard & Easy & Hard & Easy & Hard & Easy & Hard \\
    \midrule
    Average$_{\text{Horizon}=1}$
      & 61.5 & 27.0 & 85.9 & \textbf{49.7} & 76.4 & 24.8 & \textbf{91.0} & 26.7
      & \underline{90.2} & \underline{33.4} \\
    Average$_{\text{Horizon}=2}$
      & 43.6 & 6.0 & \underline{60.2} & \underline{23.6} & 28.8 & 3.2 & 55.8 & 7.0
      & \textbf{71.0} & \textbf{26.6} \\
    Average$_{\text{Horizon}=3}$
      & 32.6 & 8.6 & 44.6 & \underline{18.6} & 23.2 & 0.2 & \underline{55.8} & 6.8
      & \textbf{60.0} & \textbf{23.6} \\
    \midrule
    Average$_{20\text{ Tasks}}$
      & 49.8 & 17.2 & 69.2 & \textbf{35.4} & 51.2 & 13.3 & \underline{73.4} & 16.8
      & \textbf{77.9} & \underline{29.3} \\
    \bottomrule
  \end{tabular}}
  \vspace{-3mm}
\end{table*}

\subsection{Real-World Deployment}
\label{sec:realworld}

\begin{figure}[]
  \centering
  \includegraphics[width=\linewidth]{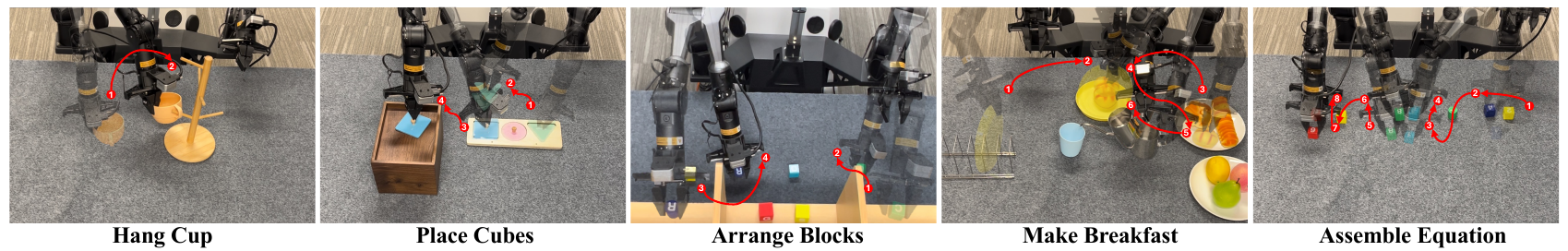}
  \caption{\textbf{Real-world tasks.} We evaluate on five manipulation tasks on an ALOHA bimanual platform: three basic tasks (Hang Cup, Arrange Blocks, Place Cubes) and two long-horizon tasks (Make Breakfast, Assemble Equation).}
  \label{fig:realworld_setup}
\end{figure}

\paragraph{Task suite.}
We deploy ThinkingVLA on an ALOHA bimanual robot platform equipped with one head camera and two wrist cameras.
We design five manipulation tasks divided into two groups (Figure~\ref{fig:realworld_setup}): three basic tasks and two long-horizon tasks.
Each task is fine-tuned with 50 demonstration episodes and evaluated over 20 trials. We report success rate (\%) per task.
More details about real-world setup and per-task descriptions are provided in Appendix~\ref{app:realworld_setup}.

\begin{figure}[]
  \centering
  \includegraphics[width=\linewidth]{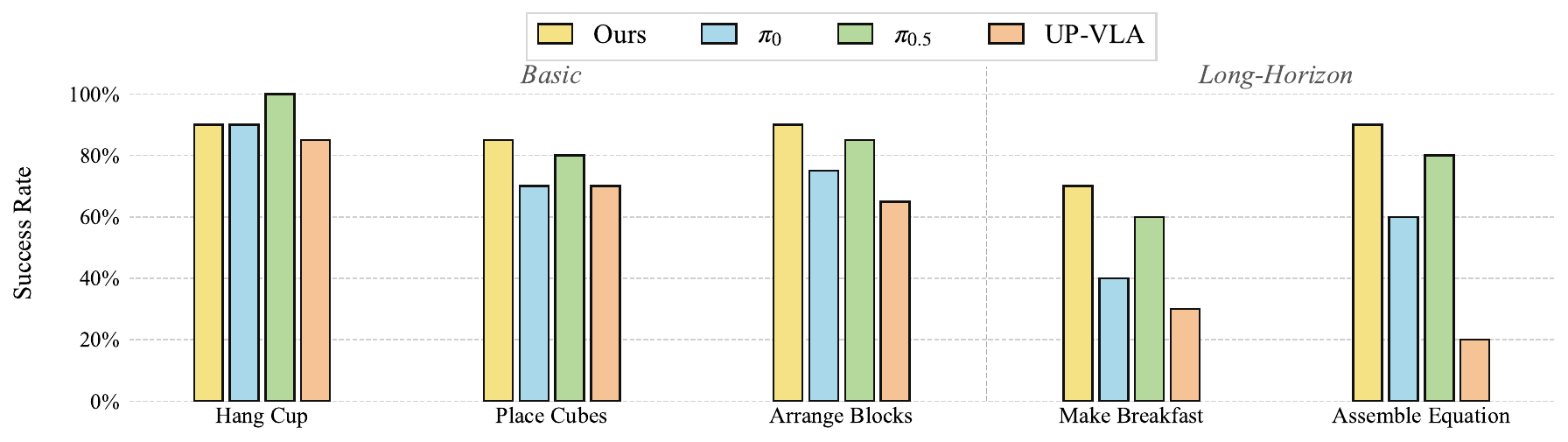}
  \caption{Real-world manipulation results on ALOHA. We report success rate (\%) over 20 trials per task. Tasks are grouped into basic (left) and long-horizon (right).}
  \label{fig:realworld_results}
\end{figure}

\paragraph{Results.}
Figure~\ref{fig:realworld_results} summarizes the real-world results.
ThinkingVLA achieves the highest success rate on four of five tasks, with the largest gains on long-horizon tasks: it outperforms $\pi_{0.5}$ by 10\,pp on both Make Breakfast and Assemble Equation. Notably, UP-VLA, which also incorporates visual prediction, drops to 30\% and 20\% on these tasks, indicating that visual prediction alone is insufficient without structured textual reasoning.
Visualizations of the reasoning process are provided in Appendix~\ref{app:qualitative}; additional robustness experiments in Appendix~\ref{app:robustness}.

\subsection{Ablation Study}
\label{sec:ablation}

We ablate reasoning components and pretraining strategy on the real-world task suite. We compare the full model against five variants:
(1)~\textbf{w/o forward CoT}: the future image is generated directly without forward reasoning;
(2)~\textbf{w/o inverse CoT}: actions are conditioned on the predicted image without inverse reasoning;
(3)~\textbf{only visual prediction}: both CoTs are removed, retaining only the predicted future image for action generation;
(4)~\textbf{only subtask}: both CoTs and visual prediction are removed, retaining only a textual subtask description;
(5)~\textbf{w/o Stage~2 pretrain}: the model fine-tunes directly from Stage~1 to Stage~3.

\paragraph{Reasoning component ablation.}
Table~\ref{tab:ablation} reports the results for variants (1)--(4).
The inverse CoT is the most critical component: removing it reduces overall success from 85.0\% to 70.0\% ($-$15\,pp), with long-horizon tasks suffering most (Make Breakfast 70$\to$45\%, Assemble Equation 90$\to$60\%).
Forward CoT provides a smaller but consistent gain of 6\,pp overall, concentrated on multi-step tasks where subgoal identification guides more accurate visual prediction.
Notably, Only visual (67.0\%) and Only subtask (69.0\%) perform comparably, both far below the full model. This shows that neither visual prediction alone nor text-only subtask planning alone is sufficient; the gain comes from their interleaved combination, where textual reasoning grounds visual prediction and the predicted image in turn grounds action reasoning.
The ablation effect scales with task complexity: on the single-step Hang Cup, all variants achieve 90\%, while on long-horizon tasks the full model outperforms the single-modality variants by 25--40\,pp.

\begin{table}[h]
  \centering
  \caption{Ablation study on real-world tasks. We report success rate (\%). Best results are \textbf{bold}.}
  \label{tab:ablation}
  \small
  \setlength{\tabcolsep}{4pt}
  \begin{tabular}{l c c c c c}
    \toprule
    \textbf{Method} & \shortstack{Hang\\Cup} & \shortstack{Arrange\\Blocks} & \shortstack{Place\\Cubes} & \shortstack{Make\\Breakfast} & \shortstack{Assemble\\Equation} \\
    \midrule
    \textbf{Full}        & \textbf{90} & \textbf{90} & \textbf{85} & \textbf{70} & \textbf{90} \\
    w/o fwd CoT          & 90 & 90 & 80 & 55 & 80 \\
    w/o inv CoT          & 90 & 80 & 75 & 45 & 60 \\
    Only visual          & 90 & 80 & 75 & 40 & 50 \\
    Only subtask          & 90 & 85 & 75 & 45 & 50 \\
    \bottomrule
  \end{tabular}
\end{table}

\begin{figure}[h]
  \centering
  \includegraphics[width=\linewidth]{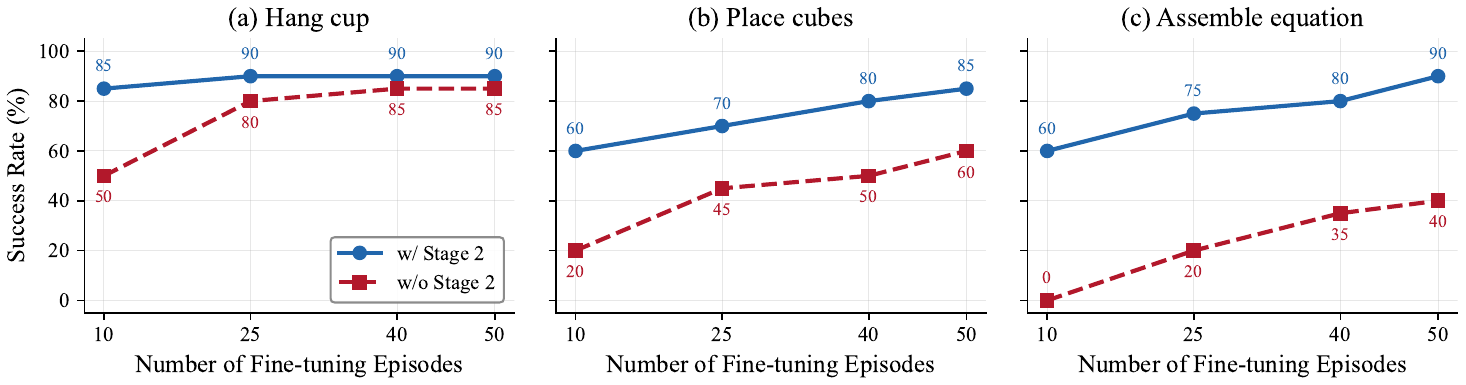}
  \caption{Effect of Stage~2 pretraining on data efficiency. We compare the full model (w/ pretrain) against a variant that skips Stage~2 (w/o pretrain), varying the number of fine-tuning episodes on three tasks: (a)~Hang Cup, (b)~Place Cubes, and (c)~Assemble Equation.}
  \label{fig:stage2_ablation}
\end{figure}

\paragraph{Effect of Stage~2 pretraining.}
To assess whether large-scale end-to-end action pretraining (Stage~2) is necessary, we compare the full model against a variant that skips Stage~2 and directly fine-tunes from Stage~1 to Stage~3. We vary the number of fine-tuning episodes (10, 25, 40, 50) on three tasks spanning different horizons and report success rate in Figure~\ref{fig:stage2_ablation}; results on the remaining two tasks are provided in Appendix~\ref{app:ablation}.
Stage~2 pretraining consistently improves data efficiency, and the benefit grows with task complexity: the gap at 50 episodes increases from 5\,pp on Hang Cup to 25\,pp on Place Cubes and 50\,pp on Assemble Equation, where the variant without pretraining fails completely at 10 episodes. These results confirm that end-to-end action pretraining provides transferable manipulation priors that cannot be recovered by additional target-domain data alone.

\section{Conclusion, Limitations and Future Work}
\label{sec:conclusion}
\vspace{-3mm}

\paragraph{Conclusion.}
We presented ThinkingVLA, a generative VLA that bridges prediction and inverse dynamics through interleaved vision--language reasoning, achieving consistent gains over unimodal and decoupled CoT approaches in both simulation and real-world settings.

\vspace{-3mm}

\paragraph{Limitations and future work.}

Generating the full reasoning chain adds latency, which is a cost inherent to all reasoning-augmented models. While reasoning dropout (\S\ref{sec:training}) already allows selective skipping, adaptive reasoning that invokes the chain only for complex or uncertain steps is a promising direction for further reducing overhead.

\clearpage

\bibliographystyle{plainnat}
\bibliography{main}

\clearpage

\clearpage
\appendix

\section{Model Architecture Details}
\label{app:architecture}

Table~\ref{tab:model_architecture} provides the full architectural specification of the released ThinkingVLA model, supplementing the overview in \S\ref{sec:architecture}.

\begin{table}[t]
  \centering
  \caption{Model architecture details for the released ThinkingVLA PyTorch configuration.}
  \label{tab:model_architecture}
  \small
  \setlength{\tabcolsep}{7pt}
  \renewcommand{\arraystretch}{1.08}
  \begin{tabular}{>{\raggedright\arraybackslash}p{0.46\linewidth}|>{\centering\arraybackslash}p{0.38\linewidth}}
    \toprule
    \textbf{Component} & \textbf{Configuration} \\
    \midrule
    \multicolumn{2}{l}{\textbf{Thinking Expert}} \\
    Backbone & PaliGemma / Gemma-2B \\
    Hidden Size & 2048 \\
    Layers & 18 \\
    Attention Heads & 8 \\
    KV Heads & 1 \\
    Head Dim. & 256 \\
    MLP Dim. & 16384 \\
    \midrule
    \multicolumn{2}{l}{\textbf{Action Expert}} \\
    Backbone & Gemma-300M \\
    Hidden Size & 1024 \\
    Layers & 18 \\
    Attention Heads & 8 \\
    KV Heads & 1 \\
    Head Dim. & 256 \\
    MLP Dim. & 4096 \\
    Action Dim. & 32 \\
    \midrule
    \multicolumn{2}{l}{\textbf{Shared Attention}} \\
    Stream Coupling & Layerwise QKV concat. \\
    Attention & Shared over prefix/suffix \\
    FFN / Norm / Output & Separate per stream \\
    Multi-view Mask & Wrist views isolated \\
    \midrule
    \multicolumn{2}{l}{\textbf{Unified Tokenizer}} \\
    Text Tokens & 257152 \\
    Special Tokens & 10 \\
    Vision Tokens & 64000 \\
    Total Vocabulary & 321162 \\
    Embedding / LM Head & Tied, 657.74M params \\
    \midrule
    \multicolumn{2}{l}{\textbf{Vision Modules}} \\
    Vision Encoder & SigLIP \\
    Encoder Hidden / Layers & 1152 / 27 \\
    Image / Patch Size & 224 / 14 \\
    Projection Dim. & 2048 \\
    Image Tokenizer & Cosmos DI16x16-360p \\
    Tokenizer Status & Frozen \\
    \midrule
    \multicolumn{2}{l}{\textbf{Action Prediction}} \\
    Input Projection & $32 \rightarrow 1024$ \\
    Output Projection & $1024 \rightarrow 32$ \\
    Time MLPs & $1024 \rightarrow 1024$ $\times$ 2 \\
    Objective & Flow matching \\
    Time Sampling & $\mathrm{Beta}(1.5, 1.0)$ \\
    Head Params & $\sim$2.17M \\
    \midrule
    \multicolumn{2}{l}{\textbf{Model Scale}} \\
    Thinking Expert & $\sim$2B \\
    Action Expert & $\sim$300M \\
    Unified Embedding / Head & 657.74M \\
    \bottomrule
  \end{tabular}
\end{table}

\section{Training Dataset Composition}
\label{app:dataset}

Stages~1 and~2 train on a mixture of 25 datasets drawn from the Open X-Embodiment collection, comprising approximately 27.3M frames of diverse robotic manipulation data across multiple robot platforms, task families, and environment configurations. Table~\ref{tab:dataset} lists each constituent dataset.

\begin{table}[t]
  \centering
  \caption{Training dataset composition for Stages~1 and~2. Proportions are computed from per-dataset frame counts (27.3M total).}
  \label{tab:dataset}
  \small
  \setlength{\tabcolsep}{5pt}
  \renewcommand{\arraystretch}{1.05}
  \begin{tabular}{lr}
    \toprule
    \textbf{Dataset} & \textbf{Mixture (\%)} \\
    \midrule
    Language Table              & 25.85 \\
    BC-Z                        & 20.08 \\
    Furniture Bench Dataset     & 14.49 \\
    Fractal                     & 13.89 \\
    Bridge                      &  6.95 \\
    Dobbe                       &  4.18 \\
    FMB                         &  4.17 \\
    Berkeley RPT                &  1.44 \\
    UTAustin Mutex              &  1.33 \\
    Stanford Hydra Dataset      &  1.31 \\
    Austin Sailor Dataset       &  1.30 \\
    Toto                        &  1.08 \\
    CMU Play Fusion             &  0.87 \\
    Taco Play                   &  0.79 \\
    IAMLab CMU Pickup Insert    &  0.54 \\
    Berkeley Autolab UR5        &  0.32 \\
    Jaco Play                   &  0.26 \\
    Viola                       &  0.25 \\
    Berkeley Fanuc Manipulation &  0.23 \\
    Berkeley MVP                &  0.17 \\
    Berkeley Cable Routing      &  0.14 \\
    NYU Franka Play Dataset     &  0.13 \\
    Austin Buds Dataset         &  0.13 \\
    CMU Stretch                 &  0.09 \\
    DLR Edan Shared Control     &  0.03 \\
    \bottomrule
  \end{tabular}
\end{table}

\section{Thinking Data Construction}
\label{app:thinking}

ThinkingVLA requires paired reasoning annotations---a forward CoT $r_t^{\mathrm{fwd}}$ and an inverse CoT $r_t^{\mathrm{inv}}$---at each training timestep. We generate these offline by prompting a multimodal language model on demonstration trajectories. For each timestep $t$, the pipeline receives three inputs: (1)~the task instruction $l$, (2)~the current observation image $o_t$, and (3)~a future observation image $o_{t+k}$, where $k$ equals the action horizon so the future frame corresponds to the ground-truth next state in the demonstration. Two separate calls produce the forward and inverse CoT respectively; both are constrained to a single paragraph of at most 60 words enclosed in \texttt{<think>...</think>} delimiters.

\paragraph{Forward CoT annotation.}
The forward CoT $r_t^{\mathrm{fwd}}$ reasons about future visual state prediction: given the current scene and task, what should the next state look like? The model is prompted with the system instruction in Table~\ref{tab:prompt_fwd} and the user message:

\begin{quote}
\small\texttt{Task: \{task\}}

\smallskip
\small\texttt{Please analyze the current observation and future observation. Generate a concise thinking process that explains how the future visual state follows from the current visual state under the task. Output your reasoning in exactly one <think>...</think> block.}
\end{quote}

\noindent followed by two images captioned \texttt{Current Observation} ($o_t$) and \texttt{Future Observation} ($o_{t+k}$). The output covers four aspects: (1)~current state analysis---positions and states of the robot arm, gripper, and target objects; (2)~task decomposition---what immediate progress is needed; (3)~future state prediction---expected visual changes from $o_t$ to $o_{t+k}$; and (4)~visual prediction details---how object positions, robot pose, and spatial relationships should change.

\paragraph{Inverse CoT annotation.}
The inverse CoT $r_t^{\mathrm{inv}}$ reasons about action inference: given the observed state transition, what immediate action primitive explains it? The model is prompted with the system instruction in Table~\ref{tab:prompt_inv} and the user message:

\begin{quote}
\small\texttt{Task: \{task\}}

\smallskip
\small\texttt{Please analyze the visual transition from the current observation to the future observation and identify the immediate action needed to advance the task. Generate your reasoning in exactly one <think>...</think> block, and include ``subtask: [action]'' inside the block.}
\end{quote}

\noindent with the same two images. The output must include an explicit \texttt{subtask:} tag identifying the inferred action primitive (e.g., \emph{approach}, \emph{grasp}, \emph{lift}, \emph{move\_to}, \emph{place}, \emph{release}, \emph{finish}), embedding a discrete action categorization within the free-form reasoning paragraph.

\paragraph{Annotation constraints.}
Both prompts enforce the following: (i)~output is a single continuous paragraph without bullet points or line breaks; (ii)~reasoning focuses on spatial relationships and visible scene elements; (iii)~no references to frame indices, timestamps, camera names, or other metadata; (iv)~\texttt{subtask: finish} is reserved for cases where the task is already fully satisfied in the current observation.

\begin{table}[t]
  \centering
  \caption{System prompt for forward CoT ($r_t^{\mathrm{fwd}}$) annotation. The placeholder \texttt{\{task\}} is replaced with the episode's task instruction at generation time.}
  \label{tab:prompt_fwd}
  \small
  \setlength{\tabcolsep}{4pt}
  \renewcommand{\arraystretch}{1.05}
  \begin{tabular}{p{0.92\linewidth}}
    \toprule
    \textbf{System Prompt --- Forward CoT} \\
    \midrule
    \textbf{Role:} You are an advanced robotic intelligence agent specializing in visual future state prediction and manipulation planning. \\[3pt]
    \textbf{Task:} Given the global task instruction, the current observation, and the future observation, generate a concise thinking process that explains how to predict the future visual state from the current visual state. \\[3pt]
    \textbf{Inputs:} (1)~Global task instruction: the overall manipulation goal. (2)~Current observation: the current visual observation of the robot, objects, and workspace. (3)~Future observation: the future visual observation after the robot has made progress toward the task. \\[3pt]
    \textbf{Required output:} Exactly one \texttt{<think>...</think>} block containing a single coherent paragraph of plain text (max 60 words), covering: current state analysis, task decomposition, future state prediction, and visual prediction details. \\[3pt]
    \textbf{Constraints:} No bullet points or line breaks. Focus on spatial reasoning. Be specific about visible object locations and robot--object relationships. No frame indices, timestamps, camera names, or metadata. \\
    \bottomrule
  \end{tabular}
\end{table}

\begin{table}[t]
  \centering
  \caption{System prompt for inverse CoT ($r_t^{\mathrm{inv}}$) annotation. The output must contain an explicit \texttt{subtask:} tag.}
  \label{tab:prompt_inv}
  \small
  \setlength{\tabcolsep}{4pt}
  \renewcommand{\arraystretch}{1.05}
  \begin{tabular}{p{0.92\linewidth}}
    \toprule
    \textbf{System Prompt --- Inverse CoT} \\
    \midrule
    \textbf{Role:} You are an advanced robotic intelligence agent specializing in visually grounded robotic manipulation planning. \\[3pt]
    \textbf{Task:} Given the global task instruction, the current observation, and the future observation, infer the immediate action primitive that best explains the transition from the current visual state to the future visual state. \\[3pt]
    \textbf{Inputs:} (1)~Global task instruction: the overall manipulation goal. (2)~Current observation: the current visual observation of the robot, objects, and workspace. (3)~Future observation: the future visual observation after an immediate task-relevant action has been taken. \\[3pt]
    \textbf{Required output:} Exactly one \texttt{<think>...</think>} block containing a single coherent paragraph of plain text (max 60 words). Must include ``\texttt{subtask: [action]}'' where \texttt{[action]} is a concise action primitive (e.g., \emph{approach}, \emph{grasp}, \emph{lift}, \emph{move\_to}, \emph{place}, \emph{release}, \emph{finish}). \\[3pt]
    \textbf{Think rules:} (1)~Spatial grounding: describe relevant robot, gripper, and object positions. (2)~State difference: identify the key physical change between observations. (3)~Immediate action: infer the single action explaining this transition. (4)~\texttt{subtask: finish} only if the goal is already fully satisfied. \\[3pt]
    \textbf{Constraints:} No bullet points or line breaks. No metadata, success/failure labels, or multiple alternative actions. \\
    \bottomrule
  \end{tabular}
\end{table}

\section{Training Hyperparameters}
\label{app:training}

ThinkingVLA is trained progressively: Stage~1 initializes from pretrained PaliGemma~\citep{beyer2024paligemma} weights; each subsequent stage initializes from the best checkpoint of the preceding stage.
Table~\ref{tab:trainable} summarizes which modules are trainable or frozen at each stage.
Tables~\ref{tab:training_stage1}--\ref{tab:training_stage3} provide the optimizer, loss-weight, and schedule configurations.

\begin{table}[t]
  \centering
  \caption{Trainable (\checkmark) and frozen (---) modules per training stage.}
  \label{tab:trainable}
  \small
  \setlength{\tabcolsep}{6pt}
  \renewcommand{\arraystretch}{1.08}
  \begin{tabular}{lccc}
    \toprule
    \textbf{Module} & \textbf{Stage 1} & \textbf{Stage 2} & \textbf{Stage 3} \\
    \midrule
    Thinking expert transformer & \checkmark & \checkmark & \checkmark \\
    Vision tower               & \checkmark & ---        & \checkmark \\
    Multimodal projector       & \checkmark & ---        & \checkmark \\
    Unified embedding / head   & \checkmark & ---        & \checkmark \\
    Action expert              & ---        & \checkmark & \checkmark \\
    Action / time heads        & ---        & \checkmark & \checkmark \\
    Cosmos tokenizer           & ---        & ---        & --- \\
    \bottomrule
  \end{tabular}
\end{table}

\begin{table}[t]
  \centering
  \caption{Stage 1 training hyperparameters.}
  \label{tab:training_stage1}
  \small
  \setlength{\tabcolsep}{6pt}
  \renewcommand{\arraystretch}{1.08}
  \begin{tabular}{>{\raggedright\arraybackslash}p{0.34\linewidth}|>{\raggedright\arraybackslash}p{0.56\linewidth}}
    \toprule
    \textbf{Setting} & \textbf{Configuration} \\
    \midrule
    Action horizon & 10 \\
    Training data & 25-dataset mix \\
    Loss weights & text $1.0$; action $0.0$; img $1.0$ \\
    Optimizer & AdamW \\
    Gradient clipping & 1.0 \\
    EMA decay & 0.999 \\
    Schedule & batch $512$; warmup $2000$; peak LR $1\mathrm{e}{-4}$; decay $30000$; decay LR $1\mathrm{e}{-5}$; steps $200000$ \\
    \midrule
    CFG drop-task prob. & 0.05 \\
    CFG drop-task-image prob. & 0.05 \\
    CFG image scale & 1.5 \\
    CFG task scale & 2 \\
    \bottomrule
  \end{tabular}
\end{table}

\begin{table}[t]
  \centering
  \caption{Stage 2 training hyperparameters.}
  \label{tab:training_stage2}
  \small
  \setlength{\tabcolsep}{6pt}
  \renewcommand{\arraystretch}{1.08}
  \begin{tabular}{>{\raggedright\arraybackslash}p{0.34\linewidth}|>{\raggedright\arraybackslash}p{0.56\linewidth}}
    \toprule
    \textbf{Setting} & \textbf{Configuration} \\
    \midrule
    Action horizon & 10 \\
    Training data & 25-dataset mix \\
    Reasoning dropout & $0.1 / 0.1$ \\
    Loss weights & text $1.0$; action $10.0$; img $2.0$ \\
    Optimizer & AdamW \\
    Gradient clipping & 1.0 \\
    EMA decay & 0.999 \\
    Schedule & batch $512$; warmup $1000$; peak LR $1\mathrm{e}{-4}$; decay $45000$; decay LR $1\mathrm{e}{-5}$; steps $100000$ \\
    \midrule
    CFG drop-task prob. & 0.05 \\
    CFG drop-task-image prob. & 0.05 \\
    CFG image scale & 1.5 \\
    CFG task scale & 2 \\
    \bottomrule
  \end{tabular}
\end{table}

\begin{table}[t]
  \centering
  \caption{Stage 3 training hyperparameters.}
  \label{tab:training_stage3}
  \small
  \setlength{\tabcolsep}{6pt}
  \renewcommand{\arraystretch}{1.08}
  \begin{tabular}{>{\raggedright\arraybackslash}p{0.34\linewidth}|>{\raggedright\arraybackslash}p{0.56\linewidth}}
    \toprule
    \textbf{Setting} & \textbf{Configuration} \\
    \midrule
    \multicolumn{2}{l}{\textbf{RoboTwin}} \\
    Action horizon & 10 \\
    Loss weights & text $1.0$; action $10.0$; img $1.0$ \\
    Optimizer & AdamW \\
    Gradient clipping & 1.0 \\
    EMA decay & 0.999 \\
    Schedule & batch $32$; warmup $2000$; peak LR $5\mathrm{e}{-5}$; decay $30000$; decay LR $5\mathrm{e}{-6}$; steps $30000$ \\
    \midrule
    \multicolumn{2}{l}{\textbf{ALOHA}} \\
    Action horizon & 50 \\
    Loss weights & text $1.0$; action $10.0$; img $1.0$ \\
    Reasoning dropout & $0.1 / 0.1$ \\
    Optimizer & AdamW \\
    Gradient clipping & 1.0 \\
    EMA decay & 0.999 \\
    Schedule & batch $64$; warmup $2000$; peak LR $1\mathrm{e}{-5}$; decay $15000$; decay LR $5\mathrm{e}{-6}$; steps $15000$ \\
    \bottomrule
  \end{tabular}
\end{table}

\section{Classifier-Free Guidance Details}
\label{app:cfg}

We describe here the training-time implementation of the classifier-free guidance (CFG) and reasoning dropout techniques introduced in \S\ref{sec:training}. Both are applied only during training; the inference-time CFG fusion formula is given in Eq.~\eqref{eq:cfg}.

\paragraph{Three-branch decomposition.}
CFG is applied exclusively to future image token generation; the forward CoT $r_t^{\mathrm{fwd}}$, inverse CoT $r_t^{\mathrm{inv}}$, and flow-matching action denoising are all produced under the full conditional. Let $f_v(o_t)$ denote the ViT tokens of the current observation, $\tilde{o}_t$ the Cosmos discrete tokens of $o_t$ (the image condition), and $l$ the task instruction. The forward CoT $r_t^{\mathrm{fwd}}$ is generated under the full condition and is shared across all three branches. Let $\hat{g}_{<k}$ denote the future image tokens generated up to position $k$. The three branches compute logits over the vision vocabulary as
\begin{align}
l_{\mathrm{base}}^k &= f_\theta\!\left(\hat{g}_{<k};\; f_v(o_t),\; r_t^{\mathrm{fwd}}\right), \label{eq:cfg_base}\\
l_{\mathrm{img}}^k  &= f_\theta\!\left(\hat{g}_{<k};\; f_v(o_t),\; \tilde{o}_t,\; r_t^{\mathrm{fwd}}\right), \label{eq:cfg_img}\\
l_{\mathrm{full}}^k &= f_\theta\!\left(\hat{g}_{<k};\; f_v(o_t),\; \tilde{o}_t,\; l,\; r_t^{\mathrm{fwd}}\right). \label{eq:cfg_full}
\end{align}
The ViT tokens $f_v(o_t)$ are retained in all three branches; the branches differ only in whether the image condition $\tilde{o}_t$ and task instruction $l$ are included. The per-token fused logit (Eq.~\eqref{eq:cfg}) thus becomes $l_{\mathrm{cfg}}^k = l_{\mathrm{base}}^k + s_I(l_{\mathrm{img}}^k - l_{\mathrm{base}}^k) + s_T(l_{\mathrm{full}}^k - l_{\mathrm{img}}^k)$, with $s_I = 1.5$ and $s_T = 2$.

\paragraph{Training branch sampling.}
Each training sample is independently assigned to one of the three branches at random:
\begin{itemize}
    \item \textbf{Full} (90\%): the complete prefix is used; all stage losses ($\mathcal{L}_{\text{text}} + \mathcal{L}_{\text{img}} + \mathcal{L}_{\text{action}}$, weighted per Eq.~\eqref{eq:total_loss}) are computed.
    \item \textbf{DropT} (5\%): the task instruction $l$ is omitted; only the image loss $\mathcal{L}_{\text{img}}$ over future image tokens $\hat{o}_{t+1}$ is computed.
    \item \textbf{DropTI} (5\%): both $l$ and $\tilde{o}_t$ are omitted; only $\mathcal{L}_{\text{img}}$ is computed.
\end{itemize}
The forward CoT $r_t^{\mathrm{fwd}}$ is retained in all branches, so CFG dropout does not interfere with thinking-loss supervision. In Stage~3, per-view attention isolation is additionally enforced within each branch's prefix to prevent cross-view token interference during multi-view image generation.

\paragraph{Reasoning dropout.}
During training, $r_t^{\mathrm{fwd}}$ and $r_t^{\mathrm{inv}}$ are each independently replaced with an empty string with probability $p_{\mathrm{drop}} = 0.1$; the structural tokens \texttt{[think]}/\texttt{[/think]} are always preserved so the sequence format remains well-defined. This trains the model to handle absent reasoning steps without encountering out-of-distribution inputs, which is orthogonal to image CFG and enables selective skipping of CoT generation at inference to reduce per-step latency.

\section{Per-Task RoboTwin 2.0 Results}
\label{app:robotwin}

Table~\ref{tab:robotwin_full} reports success rates (\%) on the 20 RoboTwin 2.0 tasks evaluated by ThinkingVLA under two conditions: \textit{Clean} (\texttt{demo\_clean}) and \textit{Rand.} (\texttt{demo\_randomized}). \textbf{Bold} marks the best result per task and condition; \underline{underline} marks the second-best.

\begin{table*}[t]
  \centering
  \caption{Per-task success rates (\%) on RoboTwin 2.0 (20 tasks, 2 conditions). Clean\,=\,\texttt{demo\_clean}; Rand.\,=\,\texttt{demo\_randomized}. \textbf{Bold}: best; \underline{underline}: second-best.}
  \label{tab:robotwin_full}
  \setlength{\tabcolsep}{4pt}
  \resizebox{\textwidth}{!}{%
  \begin{tabular}{l c cc cc cc cc cc}
    \toprule
    & & \multicolumn{2}{c}{\textbf{$\pi_0$}}
    & \multicolumn{2}{c}{\textbf{UP-VLA}}
    & \multicolumn{2}{c}{\textbf{XVLA}}
    & \multicolumn{2}{c}{\textbf{BagelVLA}}
    & \multicolumn{2}{c}{\textbf{ThinkingVLA (Ours)}} \\
    \cmidrule(lr){3-4}\cmidrule(lr){5-6}\cmidrule(lr){7-8}\cmidrule(lr){9-10}\cmidrule(lr){11-12}
    \textbf{Task} & \textbf{Hor.} & Clean & Rand. & Clean & Rand. & Clean & Rand. & Clean & Rand. & Clean & Rand. \\
    \midrule
    Adjust Bottle            & 1 &  90 &  \textbf{56} & \textbf{100} &  17 &  \underline{97} &  \textbf{56} & \textbf{100} &  14 & 92 & \underline{20} \\
    Beat Block Hammer        & 1 &  43 &  \underline{21} &  66 &  16 &  78 &  18 &  \textbf{87} &  16 & \underline{85} & \textbf{25} \\
    Click Alarmclock         & 1 &  63 &  11 &  69 &  \underline{41} &  \textbf{96} &  \textbf{69} &  85 &  20 & \underline{90} & 25 \\
    Click Bell               & 1 &  44 &   3 &  54 &  \textbf{72} & \textbf{100} &  \underline{61} & \textbf{100} &  35 & \underline{80} & 60 \\
    Dump Bin Bigbin          & 1 &  83 &  24 &  81 &  35 &  \textbf{94} &  \underline{59} &  91 &  51 & \underline{92} & \textbf{60} \\
    Grab Roller              & 1 &  \underline{96} &  \textbf{80} &  \textbf{99} &  28 &  \textbf{99} &  \underline{66} &  \textbf{99} &  41 & 95 & 38 \\
    Move Playingcard Away    & 1 &  53 &  22 &  79 &  13 &  \textbf{94} &  \textbf{57} &  \underline{92} &  \underline{30} & 90 & 28 \\
    Open Laptop              & 1 &  85 &  \underline{46} &  \underline{86} &  21 &  85 &  \textbf{73} &  \textbf{96} &  37 & \textbf{96} & 40 \\
    Place A2B Left           & 1 &  31 &   1 &  74 &   4 &  62 &  \textbf{21} &  \underline{79} &  12 & \textbf{90} & \underline{20} \\
    Place A2B Right          & 1 &  27 &   6 &  56 &   1 &  54 &  \underline{17} &  \underline{81} &  11 & \textbf{92} & \textbf{18} \\
    \midrule
    Handover Block           & 2 &  \textbf{45} &   8 &   4 &   0 &  27 &  \textbf{30} &  38 &   0 & \underline{40} & \underline{10} \\
    Handover Mic             & 2 &  \underline{98} &  13 &  45 &   0 & \textbf{100} &  \textbf{38} &  75 &   8 & 85 & \underline{28} \\
    Hanging Mug              & 2 &  11 &   \underline{3} &   4 &   0 &  \underline{34} &  \textbf{15} &  12 &   1 & \textbf{40} & \textbf{15} \\
    Place Bread Skillet      & 2 &  23 &   1 &  71 &  16 &  82 &  17 &  \underline{91} &  \underline{26} & \textbf{92} & \textbf{30} \\
    Place Can Basket         & 2 &  41 &   5 &  20 &   0 &  58 &  \underline{18} &  \underline{63} &   0 & \textbf{98} & \textbf{50} \\
    \midrule
    Blocks Ranking RGB       & 3 &  19 &   5 &  38 &   0 &  79 &  \underline{26} &  \underline{84} &   4 & \textbf{90} & \textbf{30} \\
    Blocks Ranking Size      & 3 &   7 &   1 &  21 &   0 &  42 &   \underline{9} &  \underline{45} &   2 & \textbf{50} & \textbf{15} \\
    Put Bottles Dustbin      & 3 &  \textbf{54} &  \underline{13} &   7 &   0 &   0 &   1 &  \underline{42} &  10 & 40 & \textbf{15} \\
    Stack Blocks Three       & 3 &  17 &   0 &   8 &   0 &  22 &  \underline{15} &  \textbf{45} &   5 & \underline{40} & \textbf{18} \\
    Stack Bowls Three        & 3 &  \underline{66} &  24 &  42 &   1 &  \textbf{80} &  \textbf{42} &  63 &  13 & \textbf{80} & \underline{40} \\
    \midrule
    \textbf{Average}         & $-$ &  50 &  17 &  51 &  13 &  69 &  \textbf{35} &  \underline{73} &  17 & \textbf{78} & \underline{29} \\
    \bottomrule
  \end{tabular}}
\end{table*}

\section{Real-World Setup and Task Details}
\label{app:realworld_setup}

We deploy ThinkingVLA on an ALOHA bimanual robot platform equipped with one overhead third-view camera and two wrist-mounted cameras (one per arm). All three views are used as visual observations during both training and evaluation. We design five manipulation tasks divided into three basic tasks and two long-horizon tasks; each task is trained with 50 teleoperated demonstration episodes and evaluated over 20 trials. Table~\ref{tab:task_details} summarizes the task specifications, and per-task descriptions follow.

\begin{table}[t]
  \centering
  \caption{Real-world task specifications. Horizon denotes the number of sequential subtasks required for completion.}
  \label{tab:task_details}
  \small
  \setlength{\tabcolsep}{5pt}
  \renewcommand{\arraystretch}{1.08}
  \begin{tabular}{llcl}
    \toprule
    \textbf{Task} & \textbf{Group} & \textbf{Horizon} & \textbf{Success Criterion} \\
    \midrule
    Hang Cup         & Basic       & 1 & Cup stably hanging on rack \\
    Place Cubes      & Basic       & 2 & Both cubes at target positions \\
    Arrange Blocks   & Basic       & 2 & Target letter blocks placed correctly \\
    Make Breakfast   & Long-horizon & 3 & Plate placed, bread plated, water poured \\
    Assemble Equation & Long-horizon & 3 & Correct digits placed to complete equation \\
    \bottomrule
  \end{tabular}
\end{table}

\paragraph{Hang Cup.}
The robot grasps a cup from the table and hangs it on a fixed rack. Training demonstrations vary the initial cup position across the workspace and use cups of different colors, requiring the policy to generalize over object appearance and spatial layout.

\paragraph{Place Cubes.}
Two cubes must be picked and placed at designated target positions. The training set contains three cubes of distinct shapes; each episode selects a combination of two, so the policy must identify the correct pair from the instruction and place each at the corresponding location.

\paragraph{Arrange Blocks.}
The robot picks lettered blocks and assembles them to spell ``CORL.'' Each training episode selects two of the four target blocks for grasping, so the policy must recognize which blocks to pick and where to place them to form the correct arrangement. This task tests fine-grained object discrimination and precise placement.

\paragraph{Make Breakfast.}
A long-horizon task requiring three sequential subtasks: place a plate on the table, plate a piece of bread onto the plate, and pour water from a cup. Training demonstrations use plates of varied colors, bread of different shapes, and cups of different colors, introducing appearance diversity across all three subtasks. Success requires completing the full sequence in order.

\paragraph{Assemble Equation.}
The robot assembles digit blocks to complete a given arithmetic equation (e.g., ``assemble the blocks to complete $24 - 18 = \text{?}$''). The task requires the model to first compute the correct result from the equation in the instruction, then identify and grasp the corresponding digit blocks and place them at the designated position. This task is the most demanding in our suite: it combines numerical reasoning over the language instruction with multi-step pick-and-place execution, testing whether the interleaved reasoning chain can support both semantic understanding and precise manipulation.

\section{Robustness Experiments}
\label{app:robustness}

\begin{figure}[t]
  \centering
  \includegraphics[width=\linewidth]{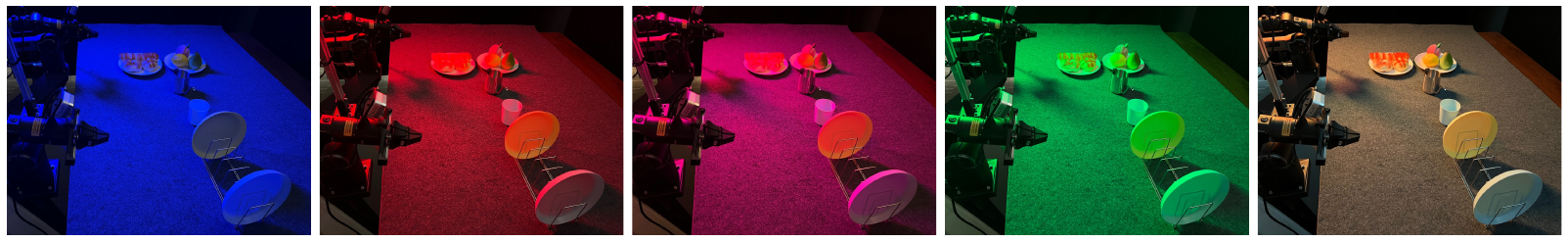}
  \caption{Lighting conditions used in the robustness evaluation. From left to right: blue, red, pink, green, and warm-white illumination.}
  \label{fig:light}
\end{figure}

To evaluate visual robustness, we test all real-world methods under varied lighting conditions that drastically alter the color distribution of the scene. We use a programmable LED light to cast uniform blue, red, pink, green, and warm-white illumination over the workspace (Figure~\ref{fig:light}), each shifting both object appearance and background color far from the training distribution. Each task is evaluated for 25 trials in total (5 trials per lighting condition); we report the aggregate success rate (\%) per task.

\begin{table}[h]
  \centering
  \caption{Robustness to lighting variation on real-world tasks. We report success rate (\%) aggregated over 25 trials per task (5 trials $\times$ 5 lighting conditions). Best results are \textbf{bold}; second-best are \underline{underlined}.}
  \label{tab:robustness}
  \small
  \setlength{\tabcolsep}{4pt}
  \renewcommand{\arraystretch}{1.08}
  \begin{tabular}{l c c c c c}
    \toprule
    \textbf{Method} & \shortstack{\textbf{Hang}\\\textbf{Cup}} & \shortstack{\textbf{Place}\\\textbf{Cubes}} & \shortstack{\textbf{Arrange}\\\textbf{Blocks}} & \shortstack{\textbf{Make}\\\textbf{Breakfast}} & \shortstack{\textbf{Assemble}\\\textbf{Equation}} \\
    \midrule
    $\pi_0$~\citep{black2024pi_0}           & 20 & 40 & 16 & 4 & 40 \\
    $\pi_{0.5}$~\citep{black2025pi_}        & \underline{40} & \underline{60} & \underline{32} & \underline{28} & \underline{56} \\
    UP-VLA~\citep{zhang2025up}               & 28 & 20 & 12 & 0 & 0 \\
    \textbf{ThinkingVLA (Ours)}              & \textbf{68} & \textbf{80} & \textbf{48} & \textbf{52} & \textbf{68} \\
    \bottomrule
  \end{tabular}
\end{table}

ThinkingVLA consistently outperforms all baselines under lighting perturbation, achieving the highest success rate on every task. Arrange Blocks and Assemble Equation are particularly affected by lighting variation, as both tasks require identifying and grasping specific blocks whose appearance changes substantially under colored illumination. All methods experience notable performance drops on these two tasks compared to standard lighting (Table~\ref{tab:ablation}). Nevertheless, ThinkingVLA retains a clear advantage (48\% and 68\% vs.\ 32\% and 56\% for the next-best method). We attribute this in part to the robustness of the visual prediction module: because the image generation model is trained with data augmentation, the predicted future image is less sensitive to input color shifts than the raw observation, effectively providing a more stable visual context for downstream action reasoning. UP-VLA, which also incorporates visual prediction, does not exhibit the same robustness, suggesting that the interleaved reasoning structure, in which textual CoT mediates between observation and image generation, may play a role in stabilizing the visual prediction under distribution shift.

\section{Additional Ablation Studies}
\label{app:ablation}

\begin{figure}[t]
  \centering
  \includegraphics[width=\linewidth]{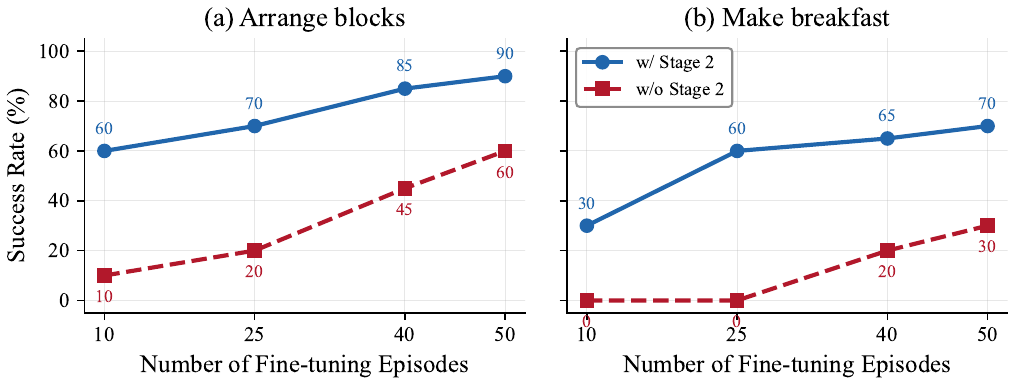}
  \caption{Effect of Stage~2 pretraining on the remaining two real-world tasks: (a)~Arrange Blocks and (b)~Make Breakfast. Format follows Figure~\ref{fig:stage2_ablation}.}
  \label{fig:stage2_appendix}
\end{figure}

Figure~\ref{fig:stage2_appendix} reports the Stage~2 pretraining ablation on the two tasks not shown in Figure~\ref{fig:stage2_ablation}: Arrange Blocks (Horizon$=$2) and Make Breakfast (Horizon$=$3). Stage~2 pretraining improves data efficiency on both tasks, and the benefit is larger for the longer-horizon task, consistent with the main results. On Make Breakfast, the variant without pretraining fails completely at both 10 and 25 episodes (0\%) and reaches only 30\% at 50 episodes, whereas the pretrained model achieves 70\%---a 40\,pp gap. On Arrange Blocks, the gap is largest at 10 episodes (60\% vs.\ 10\%) and narrows to 30\,pp at 50 episodes as more demonstrations compensate for the missing priors. Together with the main-text results, the gaps at 50 episodes increase monotonically with task horizon: 5\,pp (Hang Cup, H$=$1) $\to$ 25--30\,pp (Place Cubes and Arrange Blocks, H$=$2) $\to$ 40--50\,pp (Make Breakfast and Assemble Equation, H$=$3), confirming that end-to-end action pretraining is most critical for complex, multi-step tasks.

\section{Real-World Reasoning Visualization and Prediction Quality}
\label{app:qualitative}

\begin{figure}[t]
  \centering
  \includegraphics[width=\linewidth]{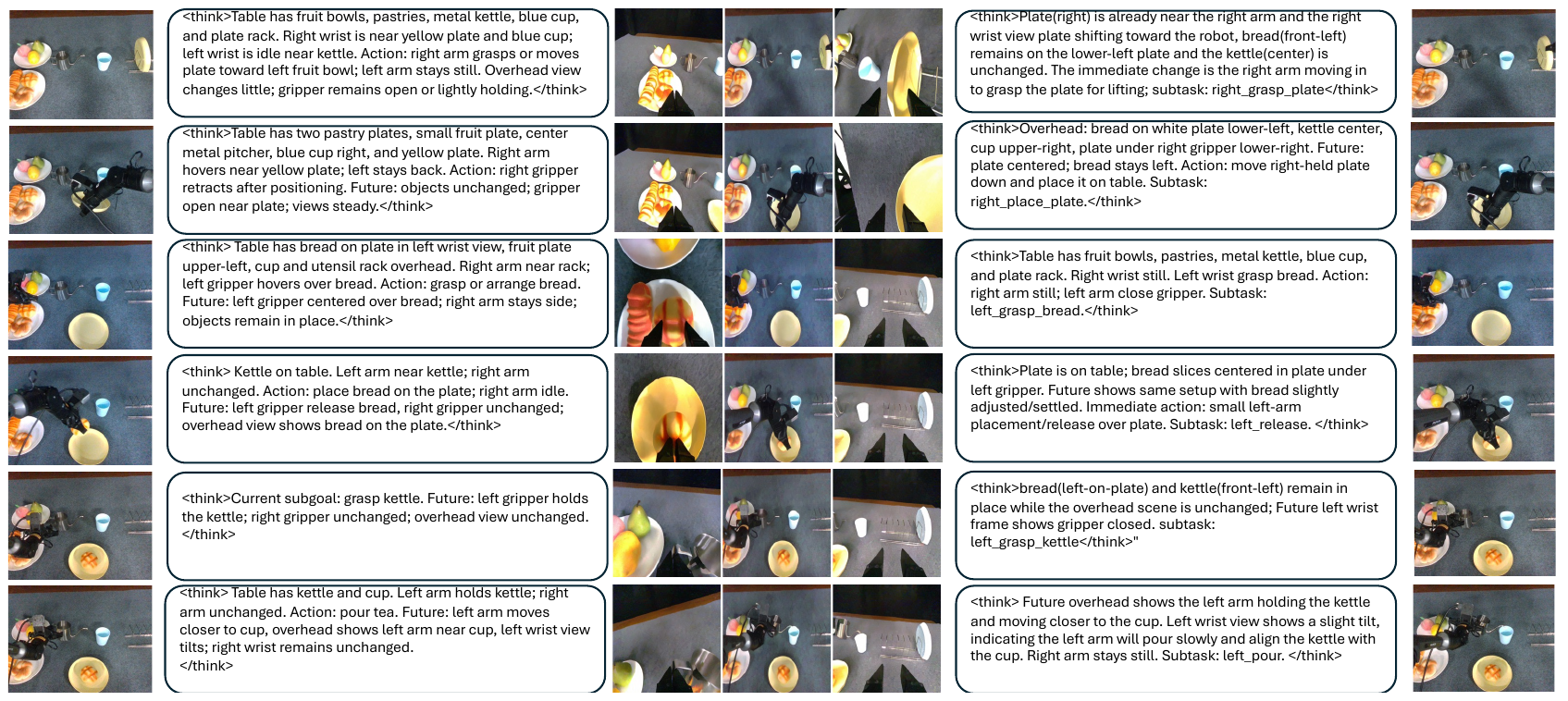}
  \caption{\textbf{Interleaved reasoning chain on Make Breakfast} (Horizon$=$3). Each row shows one decision step: the current observation (left), forward CoT reasoning about the expected visual change (black box), predicted multi-view future images (center), and inverse CoT identifying the action primitive (black box, right). Six steps cover the full task: grasp plate $\to$ place plate $\to$ grasp bread $\to$ place bread $\to$ grasp kettle $\to$ pour water. The predicted images closely match the actual observations at the next step, confirming accurate visual forecasting.}
  \label{fig:qual_breakfast}
\end{figure}

We visualize the complete interleaved reasoning chain produced by ThinkingVLA on a successful Make Breakfast rollout (Figure~\ref{fig:qual_breakfast}). This long-horizon task requires three sequential subtasks---grasp and place a plate, plate bread onto it, and pour water from a kettle---and thus exercises the full reasoning pipeline across six decision steps.

Each row in Figure~\ref{fig:qual_breakfast} corresponds to one reasoning step and shows, from left to right: (1)~the current multi-view observation $o_t$ (third-view), (2)~the forward CoT $r_t^{\mathrm{fwd}}$ that analyzes the scene and predicts the expected visual change, (3)~the predicted multi-view future images $\hat{o}_{t+1}$ (third-view, left wrist, right wrist), and (4)~the inverse CoT $r_t^{\mathrm{inv}}$ that identifies the immediate action primitive (e.g., \texttt{left\_grasp\_bread}, \texttt{left\_release}, \texttt{left\_pour}). Comparing the predicted images with the actual observations in the subsequent step confirms that the visual prediction module produces spatially faithful forecasts: object positions, gripper states, and scene layouts are accurately anticipated, providing reliable visual context for inverse dynamics reasoning.

The reasoning chain exhibits two notable properties. First, the forward CoT adapts its focus as the task progresses: early steps reason about object locations and grasping targets, while later steps track multi-object relationships (e.g., bread-on-plate, kettle-near-cup). Second, the inverse CoT consistently grounds its action decisions in the predicted visual state, referencing specific spatial cues (gripper closure, arm proximity, pouring angle) rather than relying on abstract subtask labels alone.

\begin{figure}[t]
  \centering
  \includegraphics[width=\linewidth]{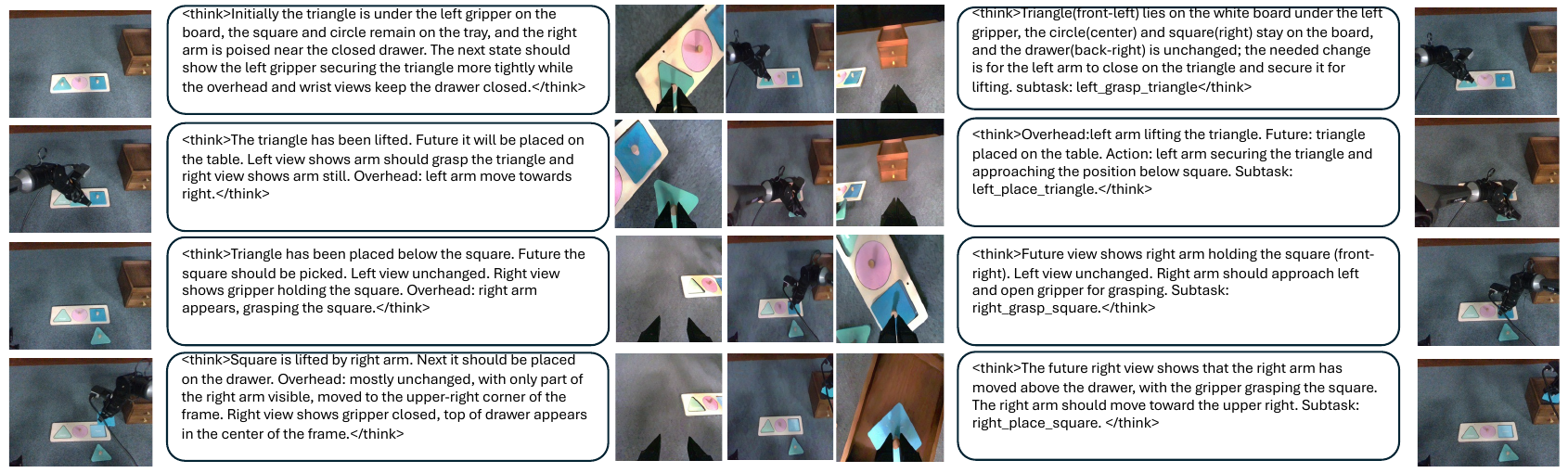}
  \caption{\textbf{Interleaved reasoning chain on Place Cubes} (Horizon$=$2)}
  \label{fig:qual_cubes}
\end{figure}

The visualization of a successful Place Cubes rollout is shown in Figure~\ref{fig:qual_cubes}.

\end{document}